\documentclass[sigconf]{acmart}

\usepackage{multirow}
\usepackage{subcaption}
\usepackage{float}
\usepackage[ruled,vlined,linesnumbered]{algorithm2e}
\usepackage{url}
\usepackage{hyperref}

\AtBeginDocument{%
  \providecommand\BibTeX{{%
    \normalfont B\kern-0.5em{\scshape i\kern-0.25em b}\kern-0.8em\TeX}}}

\copyrightyear{2020}
\acmYear{2020}
\setcopyright{acmcopyright}\acmConference[preprint]{}
\acmDOI{0000/00000}

\begin{document}

\title{Fatigue Assessment Using ECG and Actigraphy Sensors}

\author{Yang Bai}
\affiliation{%
  \institution{Open Lab, School of Computing,}
  \city{Newcastle University, UK}
}
\email{y.bai13@newcastle.ac.uk}

\author{Yu Guan}
\affiliation{%
  \institution{Open Lab, School of Computing,}
  \city{Newcastle University, UK}
}
\email{yu.guan@newcastle.ac.uk}

\author{Wan-Fai Ng}
\affiliation{%
  \institution{Translational and Clinical Research Institute, Newcastle University, UK}
}
\email{wan-fai.ng@newcastle.ac.uk}

\renewcommand{\shortauthors}{Bai et al.}

\begin{abstract}
Fatigue is one of the key factors in the loss of work efficiency and health-related quality of life, and most fatigue assessment methods were based on self-reporting, which may suffer from many factors such as recall bias.     
To address this issue, we developed an automated system using wearable sensing and machine learning techniques for objective fatigue assessment. 
ECG/Actigraphy data were collected from subjects in free-living environments. 
Preprocessing and feature engineering methods were applied, before interpretable solution and deep learning solution were introduced.
Specifically, for interpretable solution, we proposed a feature selection approach which can select less correlated and high informative features for better understanding system's decision-making process.
For deep learning solution, we used state-of-the-art self-attention model, based on which we further proposed a consistency self-attention (CSA) mechanism for fatigue assessment. 
Extensive experiments were conducted, and very promising results were achieved. 

\end{abstract}




\maketitle

\section{Introduction}

Fatigue is one of the main medical symptoms to define weakness and ageing problems \citep{agingfatigue}, and in many chronic diseases it is also a key factor in the loss of work efficiency and health-related quality of life (HRQoL), which may impose considerable health and economic burdens \citep{NHSfatigue}.
Reliable and sensitive fatigue assessment is a key aspect of evaluation of the therapeutic effects of treatments. 
\begin{figure}[H]
 \centering
 \includegraphics[width=0.7\columnwidth]{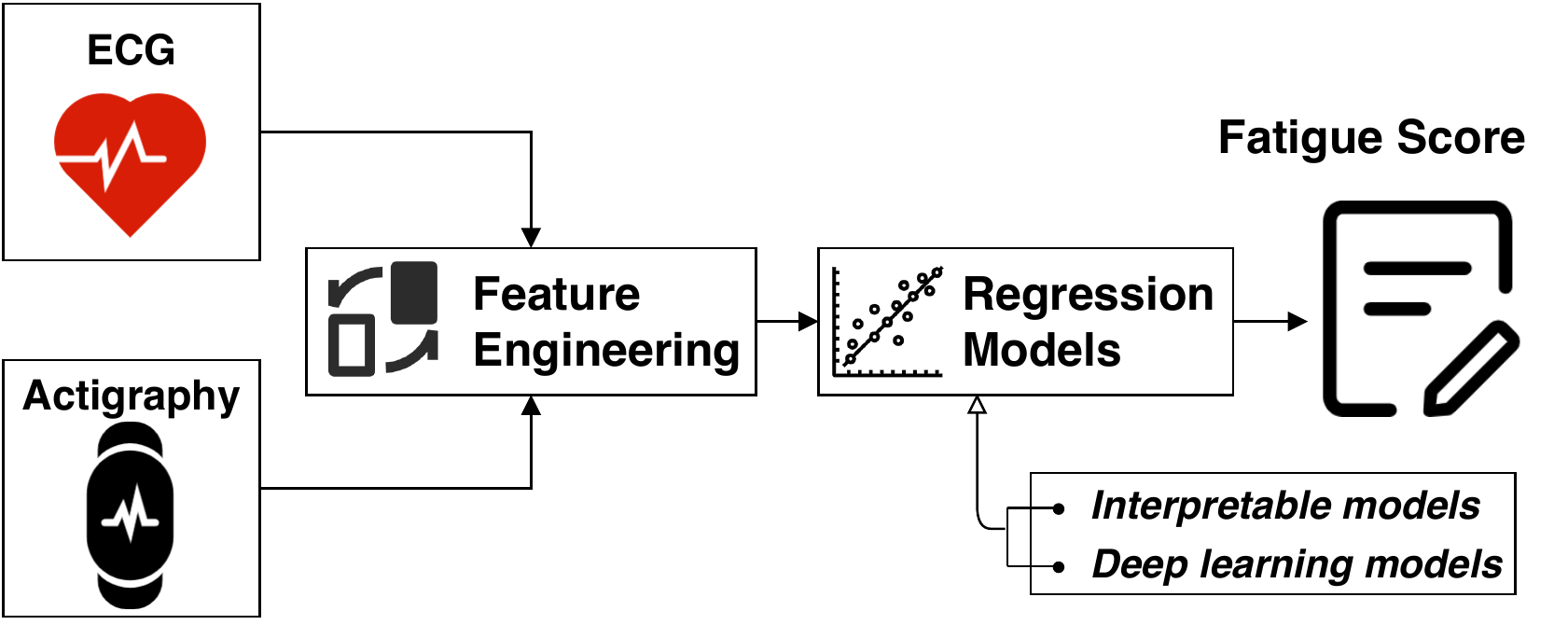}  
 \caption{An overview of our fatigue assessment system}
 \label{fig:overview}
\end{figure}
Most fatigue assessments are based on self-reporting \citep{fatigue2003psychometric}. 
However, like most questionnaire-based approaches, such subjective measurement has key limitations such as recall bias \cite{recall_bias}, and is challenging to quantify in a repeatable and reproducible way \cite{hardrepeat} \cite{aryal2017monitoring}. 
Although its reliability 
can be improved via tracking in long-terms or with high-frequency during short periods, the costs and patient burden can be increased significantly.   
Recently, sensing and machine learning (ML) techniques have been widely used for automated health assessment \cite{Shane_iswc, autism, speech_health, Tang_stroke, PD_1, PD_2, baby_stroke}. 
Through modelling the collected behaviour or physiological signals, health can be assessed in an objective and continuous manner, in contrast to the subjective, and non-continuous self-reporting methods. 
For automated fatigue assessment, potential sensing modalities include accelerometer\cite{Physical_fatigue_Jerk, physical_fatigue_HR_acc_Lasso}, HR\cite{fatigue_PPG, physical_fatigue_HR_acc_Lasso}, ECG\cite{mental_ECG}, EEG\cite{eeg2008}, EMG\cite{Physical_fatigue_muscle_exhaustion}, etc. 
based on which ML methods were used for modelling. 
Yet these approaches were tested on subjects in controlled environments.

In this work, we developed models based on ECG/Actigraphy data collected from free-living environments. 
The participants were told to wear Actigraph watch and Vital Patch for 7 days, and they were required to self-report fatigue levels 4 times a day.
After preprocessing and feature engineering, we developed regression systems to map the signals into the fatigue score for objective fatigue assessment, as shown in Fig. \ref{fig:overview}.   
Specifically, both interpretable solution and deep learning solution were proposed for modelling.

\section{Methodology}
In this paper, we aim to develop a fatigue assessment system, which can automatically output fatigue scores given Actigraphy/ECG.  
Fig. \ref{fig:overview} demonstrates an overview of this system, and in this section details are provided from data collection, ECG/Actigraphy data processing, to interpretable and deep learning solutions.

\subsection{Data Collection} \label{sec:data_collection}
In this preliminary study, 13 participants were recruited and data was collected to cover two clinical visits. 
After each visit, the participants were required to wear two medical-grade devices (for 7 days in free-living environments), namely, Actigraph GT9X Link\cite{ActiGraph} and Vital Patch\cite{vitalpatch}, from which Actigraphy/ECG data can be obtained.
The visual analog scale questionnaires \cite{seror2011eular} were also sent out to participants asking the question "How severe has your fatigue been now?". 
The fatigue score ranging from 0 to 10 represents fatigue levels from "No fatigue"(0) to "Maximal imaginable fatigue"(10).
During the 7-day data collection period, the subjects were asked to record their fatigue levels 4 times a day, i.e., morning(6am-12pm), afternoon(12pm-6pm), evening(6pm-12am), and night(12am-6am).
Accordingly, we divided daily ECG/Actigraphy signals into 4 segments, corresponding to the 4 daily recorded fatigue scores.


\subsection{Preprocessing and Feature Engineering}\label{sec:preprocessing}
After ECG/Actigraphy collection it is crucial to filter out the irrelevant information, caused by various reasons such as device artifacts, inappropriate wearing positions, non-wearing, etc. 
After that, modality-specific feature engineering can be performed for compact representation.

\subsubsection{Processing ECG}\hfill\\
ECG signal can be distorted due to various artifacts, e.g., equipment inducing movement artifacts \cite{ECG_artifacts}. 
Heart rate variability (HRV), derived from ECG, was normally considered as a reliable measure that is less sensitive to these artifacts \cite{HRV_review}. 
HRV features can be extracted based on Normal-Normal Interval (NNI), which can be achieved by R-R interval(RRI) detection, followed by a post-processing. 
In this work, we segmented the raw ECG into 5-min windows, and for each window to estimate NNI we performed the following procedure:
\begin{enumerate}
\item detecting the R-points/R-peaks (in QRS complex); 
\item computing the RRIs based on the detected R-points;
\item removing RRI outliers, i.e., the ones with the range 300-2000 ms \cite{tanaka2001age} and the ones with a difference by more than 20\% than the previous interval;
\item linearly interpolating the removed RRIs; 
\end{enumerate}
We achieved Step(1) using Python package NeuroKit\cite{neurokit} and step(3)-(4) by using toolbox HRV-Analysis\cite{hrv_analysis}.
Based on NNIs, we further assessed the data quality for each window and removed the ones with low quality. 
Given NNIs for each window, based on the source code \footnote{\url{https://github.com/bzhai/multimodal_sleep_stage_benchmark/blob/master/notebooks/Tutorial-HRV\%20Feature\%20Extraction\%20From\%20ECG.ipynb}} provided by \cite{HRV_sleep} we extracted \textbf{30 HRV features}
in four domains (time, geometrical, frequency, and non-linear domains), as shown in Table \ref{tab:featureHRV}.

\begin{table}[]
\caption{HRV features \cite{HRV_sleep}}
\scriptsize
\begin{tabular}{|l|l|}
\hline
Feature Name               & Explanation                                                                                                                                                                  \\ \hline
\multicolumn{2}{|c|}{Time Domain Features}                                                                                                                                                                \\ \hline
minimum HR                 & minimum heart rate for the interval                                                                                                                                          \\ \hline
maximum HR                 & maximum heart rate for the interval                                                                                                                                          \\ \hline
mean HR                    & mean heart rate for the interval                                                                                                                                             \\ \hline
Std HR                     & standard deviation heart rate for the interval                                                                                                                               \\ \hline
SDSD                       & Standard deviation of NNI differences                                                                                                                                        \\ \hline
SDNN                       & Standard deviation of NNI (Normal-to-Normal interval)                                                                                                                        \\ \hline
NN.mean                    & mean of normal to normal interval                                                                                                                                            \\ \hline
NN20                       & Number of successive NNI pairs that differ more than 20 ms                                                                                                                   \\ \hline
NN50                       & Number of successive NNI pairs that differ more than 50 ms                                                                                                                   \\ \hline
PNN50                      & NN50 divided by the total number of NN intervals                                                                                                                             \\ \hline
PNN20                      & NN20 divided by the total number of NN intervals                                                                                                                             \\ \hline
rMSSD                      & Square root of the mean squared differences between successive NNI                                                                                                           \\ \hline
median NN intervals        & median of NNI                                                                                                                                                                \\ \hline
range NN intervals         & range between smallest NNI to largest NNI                                                                                                                                    \\ \hline
CVSD                       & \begin{tabular}[c]{@{}l@{}}the coefficient of variation of   successive differences, the RMSSD \\ divided by mean NN\end{tabular}                                            \\ \hline
Coeff. Of Variation of NNI & the coefficient of variation of NNI                                                                                                                                          \\ \hline
\multicolumn{2}{|c|}{Frequency Domain}                                                                                                                                                                    \\ \hline
heart rate PSD             & power spectral density                                                                                                                                                       \\ \hline
low frequency (LF)         & \begin{tabular}[c]{@{}l@{}}variance (i.e. power) in HRV in the low frequency (0.04 to 0.15HZ) \\ Reflects a mixture of sympathetic and parasympathetic activity\end{tabular} \\ \hline
high frequency (HF)        & \begin{tabular}[c]{@{}l@{}}variance (i.e. power)   in HRV in the high frequency (0.15 to 0.40HZ). \\ Reflects fast changes in beat-to-beat variability\end{tabular}          \\ \hline
very low frequency (VLF)   & variance (i.e. power)   in HRV in the low frequency (0.003 to 0.04HZ)                                                                                                        \\ \hline
LF/HF                      & Ratio between LF and HF band powers                                                                                                                                          \\ \hline
Norm. low freq. Ratio      &        {LF divided by the total spectral power}                                                                                                                                                                      \\ \hline
Norm. high freq. Ratio     &       {HF divided by the total spectral power}                                                                                                                                                                      \\ \hline
\multicolumn{2}{|c|}{Non-Linear Domain}                                                                                                                                                                   \\ \hline
Cardiac Sympathetic IdNx   & Cardiac Sympathetic Index\cite{ponnusamy2012comparison}                                                                                                                                                      \\ \hline
Mod. Cardiac Symp. IdNx    & a modified cardiac sympathetic index calculated by (sd2)\textasciicircum{}2/sd1                                                                                              \\ \hline
sd1                        & Poincaré plot standard deviation perpendicular the line of   identity                                                                                                        \\ \hline
sd2                        & Poincaré plot standard deviation along the line of   identity                                                                                                                \\ \hline
sd1/sd2                    & Ratio of SD1 to SD2                                                                                                                                                          \\ \hline
cardiac vagal IndeNx       & Cardiac Vagal   IndeNx\cite{ponnusamy2012comparison}                                                                                                                                                       \\ \hline
\multicolumn{2}{|c|}{Geometrical   Domain}                                                                                                                                                                \\ \hline
Triangular Index           & The integral of the density distribution                                                                                                                                     \\ \hline
\end{tabular}
\label{tab:featureHRV}
\end{table}

\subsubsection{Processing Actigraphy}\hfill\\
\label{sec:processing_actigraphy}
For Actigraphy, one major quality issue is the missing data problem caused by non-wearing.
Via ActiLife\cite{Actilife}, we managed to calculate the Actigraphy counts every 30 seconds, and detect the non-wear time (for invalid data removal).
Within every 5-min window, based on Actigraphy counts we further extracted \textbf{8 statistical features}, i.e.,mean, median, standard deviation, variance, minimum value, maximum value, skewness and kurtosis for further processing.


\subsubsection{Multimodal Feature Sequence Construction}\hfill\\
\label{sec:data_quality}
Since there exists substantial missing data for both modalities (ECG/Actigraphy), we only preserved the ones (i.e., 5-min windows) when both are valid. 
In the segment-level (i.e., up to 6 hours), we further excluded the ones with less than 100 minutes of usable data.

After preprocessing and feature engineering, the original segment can be transformed into a $D$-dimensional sequence $\boldsymbol{X} = \{\boldsymbol{x}_t\in \mathbb{R}^D \}^T_{t=1}$, where $T$ is the sequence length (i.e., the number of windows/epoches within a segment), and $\boldsymbol{x}_t$ may be referred to as features extracted from Actigraphy, ECG, or both at timestamp $t$ (i.e., the $t^{th}$ window/epoch in a segment).
It is worth noting that due to the aforementioned data cleaning operations,
the sequence length $T$ is a non-fixed number, and in our dataset we have $T$'s
mean $\pm$ std: $56.9\pm 13.6$ with maximum/minimum value 72(i.e.,6 hours) and 21 (i.e.,105 minutes), respectively.


\subsection{Interpretable Solution}\label{sec:interpretable}
In this subsection, we aim to extract the interpretable features from sequence $\boldsymbol{X} = \{\boldsymbol{x}_t\in \mathbb{R}^D \}^T_{t=1}$, before mapping them to the corresponding fatigue score $y \in \{0,1,...,10\}$
 

\subsubsection{High-level Feature Extraction}\hfill\\
There are several interpretable machine learning models such as linear models, decision tree, etc., yet they cannot be directly applied to time series data. 
To address this issue, we further extracted high-level features from the sequence $\boldsymbol{X}$ over the time axis.
Specifically, for each dimension (out of $D$), we calculated \textbf{11 descriptive statistics}, namely 10th percentile, 25th percentile, 50th percentile, 75th percentile, 90th percentile, mean, minimum, maximum, standard deviation, skewness and kurtosis.
The high-level feature vector $\boldsymbol{x}^{high}\in \mathbb{R}^{11D}$ can be formed simply by vector concatenation.


\subsubsection{Feature Selection}\hfill\\
For $\boldsymbol{x}^{high}\in \mathbb{R}^{11D}$, there may exist high-level of feature redundancy.
For example, in Fig. \ref{fig:featCorr} we visualised the feature correlation matrix corresponding to the  combined ECG/Actigraphy modality, and we can observe high-level of feature correlation (as indicated by brighter colours). 
Although there exists feature decorrelation and dimension reduction algorithms such as PCA, in order to preserve the interpretability of the features 
here we proposed to use a Feature Selection(FS) mechanism.  
Given training set $\{\boldsymbol{X}^{high}, \boldsymbol{y}\}$, where $\boldsymbol{x}^{high} \in \boldsymbol{X}^{high}, y \in \boldsymbol{y}$, we performed the following procedure: 
\begin{enumerate}
\item similar to Fig.\ref{fig:featCorr}, calculating the correlations among all features, i.e., $corr(\boldsymbol{X}^{high},\boldsymbol{X}^{high})$;
\item grouping features together with pair-wise correlation coefficients higher than a threshold ($0.8$ in this work);
\item calculating the correlation coefficients between each feature and the fatigue score, i.e., $corr(\boldsymbol{X}^{high},\boldsymbol{y})$;
\item selecting the top feature from each group based on the coefficients  $corr(\boldsymbol{X}^{high},\boldsymbol{y})$;
\item performing LASSO to further select the key features.
\end{enumerate}
It is worth noting owing to high dimensionality, before applying LASSO we performed the "coarse feature selection" mechanism via step(2)-(4), which can select the less correlated features with highest relevance (to fatigue score).
LASSO, a data-driven feature selection approach, can further refine the feature selection procedure. 
Based on the selected features, interpretable regressors (e.g., linear models, tree-based approaches) can then be used. 

\subsection{Deep Learning Solution}
In interpretable solution, during high-level feature extraction only 11 simple statistical features were used to encode the time-series, which may cause information loss. 
A popular way to preserve the temporal information is to use deep sequential modelling such as long short term memory(LSTM) \cite{lstm}, which can encode input sequence $\boldsymbol{X}= \{\boldsymbol{x}_t\in \mathbb{R}^D \}^T_{t=1}$ into hidden states $\boldsymbol{H} = \{\boldsymbol{h}_t\in \mathbb{R}^{D_l} \}^T_{t=1}$, before using the last hidden state $\boldsymbol{h}_T$ for prediction. Note $D_l$ is the hidden state dimension.
LSTM can be trained by back-propagation through time (BPTT) \cite{lstm}, and for regression problem, mean squared error (MSE) loss $\mathcal{L}_{MSE}$ is normally used. 

\subsubsection{LSTM with Self-Attention (LSTM-SA)}\hfill\\
Compared with LSTM which uses the last hidden state $\boldsymbol{h}_{T}$ for prediction, Self-Attention(SA) \cite{transformer} further employs information from the sequence by utilising an attention vector $\boldsymbol{\alpha}\in  \mathbb{R}^T$. 
In LSTM-SA model, three additional model parameters are needed to be estimated, namely $\boldsymbol{W}^Q, \boldsymbol{W}^K, \boldsymbol{W}^V \in \mathbb{R}^{D_{a} \times D_l}$, where $D_a$ is the attention dimension.
Given the hidden states $\boldsymbol{H} \in \mathbb{R}^{D_l \times T}$ (from LSTM), three matrices named Queries $\boldsymbol{Q}\in \mathbb{R}^{D_{a} \times T}$, Keys $\boldsymbol{K}\in \mathbb{R}^{D_{a} \times T}$, and Values $\boldsymbol{V}\in \mathbb{R}^{D_{a} \times T}$ can be calculated via linear transformations such that $\boldsymbol{K} = \boldsymbol{W}^K\boldsymbol{H}$, 
$\boldsymbol{Q} = \boldsymbol{W}^Q\boldsymbol{H}$,
and $\boldsymbol{V} = \boldsymbol{W}^V\boldsymbol{H}$. 
Specifically, $\boldsymbol{K}$ is employed to learn the distribution of attention matrix on condition of $\boldsymbol{Q}$. $\boldsymbol{V} $ is used to exploit information representation.
Given that, attention vector can be calculated via 
\begin{equation}
      \boldsymbol{\alpha} = softmax(\frac{\boldsymbol{K^T} \boldsymbol{q}_T}{\sqrt{D_a}}),
      \label{eq:att weights}
\end{equation}
where $\boldsymbol{q}_T$ is the $T^{th}$ column of $\boldsymbol{Q}$.  
With attention vector $\boldsymbol{\alpha}$, the new $D_a$-dimensional representation $\boldsymbol{z}_T = \boldsymbol{V}\boldsymbol{\alpha}$ can be used for prediction.
For regression problems, 
$\mathcal{L}_{MSE}$ can be used. 

\subsubsection{LSTM with Consistency Self-Attention (LSTM-CSA)}\hfill\\
LSTM-SA is a powerful tool in many applications, but it has a non-smooth attention distribution $\boldsymbol{\alpha}$  over the sequence.
For continuous signals, temporal attention regularisation was normally used which encourages the continuity \cite{zeng2018understanding}.
Although our sequence may not be strictly continuous (due to feature engineering or missing data), there may exist certain levels of consistency for the adjacent entries, and thus we used the following regularisation term:
\begin{equation}
\Omega(\boldsymbol{\alpha})\ = T\sum_{t}{\left|\alpha_t-\ \alpha_{t-1}\right|\ },
\label{eq:regularisation}
\end{equation}
where $\alpha_t \in \boldsymbol{\alpha}$. 
The loss function can then be updated to $\mathcal{L} = \mathcal{L}_{MSE}+\lambda \Omega(\boldsymbol{\alpha})$, where $\lambda$ is the regularisation parameter.
It is worth noting that with $T$ in Eq. (\ref{eq:regularisation}), $\Omega(\boldsymbol{\alpha})$ tends to penalise heavily with a larger $T$ (i.e., with less or no missing data) to maintain its global consistency (i.e.,continuity). 

\section{Experiments}
The collected data suffered from quality issues from various reasons, 
resulting in a substantially reduced data size after data preprocessing.
Data used in the experiments were from 9 subjects, with 198 sequences. 
The demographic information of these subjects can be found in Table \ref{tab:pinfo} (in Appendix).
To evaluate the prediction models, unless stated otherwise we performed 5-fold cross-validation (5-fold CV). MAE/RMSE were used as the evaluation metrics. 

The implementation details of our methods can be found at: 
\url{https://github.com/baiyang4/Sjogrens\_questionnaire}


\subsection{Evaluation Results}
\subsubsection{Interpretable Models}\hfill\\
Based on the high-level features (e.g.,$\boldsymbol{x}^{high}$ in Sec. \ref{sec:interpretable}) and our Feature Selection (FS) method, linear regression was used for prediction.
In Table \ref{tab:results_LR}, we reported results based on different settings, from which we can observe that the performance deteriorates significantly without FS (for all modalities).
In terms of sensing modality we see Actigraphy has the worse results. 
One of the possible explanations is the inadequate feature engineering---only 8 simple statistical features were extracted from Actigraphy (see Sec. \ref{sec:processing_actigraphy}).
In contrast, we extracted the domain knowledge-driven features from ECG, yielding much better results.

\begin{table}[H]
\caption{Results of linear regression models}
\begin{tabular}{|l|l|l|l|}
\hline
Modalities                                                                    & Feat. Select. & MAE             & RMSE           \\ \hline
\multirow{3}{*}{Actigraphy}                                                & No (\#Feat. 88)            & 3.42$\pm$0.91       & 5.21$\pm$2.95      \\ \cline{2-4} 
                                                                           & Yes (\#Feat. 6)             & 2.56$\pm$0.21       & 3.10$\pm$0.22      \\ \cline{2-4} 
                                                                           & Yes (\#Feat. 3)             & 2.58$\pm$0.27       & 3.09$\pm$0.28      \\ \hline
\multirow{3}{*}{ECG}                                                       & No (\#Feat. 330)             & 4.12$\pm$0.38       & 5.26$\pm$0.47      \\ \cline{2-4} 
                                                                           & Yes (\#Feat. 30)             & \textbf{1.98$\pm$0.26}       & \textbf{2.46$\pm$0.29}      \\ \cline{2-4} 
                                                                           & Yes (\#Feat. 15)             & 2.13$\pm$0.16       & 2.57$\pm$0.25      \\ \hline
\multirow{3}{*}{\begin{tabular}[c]{@{}l@{}}ECG+\\ Actigraphy\end{tabular}} & No (\#Feat. 418)            & 4.72$\pm$0.39       & 6.01$\pm$0.51      \\ \cline{2-4} 
                                                                           & Yes (\#Feat. 30)             & 2.14$\pm$0.12       & 2.60$\pm$0.17      \\ \cline{2-4} 
                                                                           & Yes (\#Feat. 15)             & \textbf{2.11$\pm$0.11}       & \textbf{2.56$\pm$0.16}      \\ \hline
\end{tabular}
\label{tab:results_LR}
\end{table}

\begin{figure}[]
 \centering
 \begin{subfigure}[b]{0.6\columnwidth}
        \centering
        \includegraphics[width=0.95\textwidth]{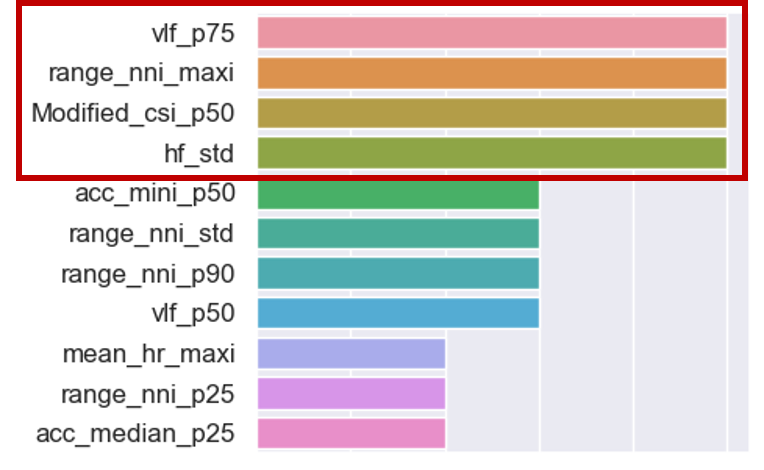}
        \caption{Feature Importance}
        \label{fig:feat_importance}
    \end{subfigure}%
    \begin{subfigure}[b]{0.4\columnwidth}
        \centering
         \includegraphics[width=0.95\textwidth]{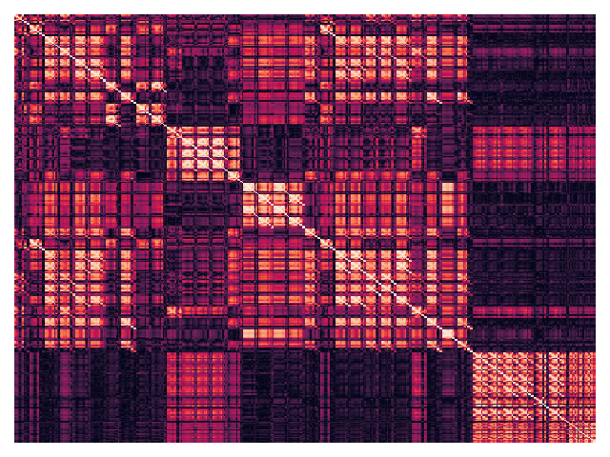}  
          \caption{Feature correlation matrix (ECG+Actigraphy)}
          \label{fig:featCorr}
    \end{subfigure}%
 \caption{(a) Red box indicating the most important features}
\end{figure}

One of the advantages of FS is to rank the most important features for better interpretation. 
In Fig. \ref{fig:feat_importance}, we demonstrated some of the most important features, which were the aggregate from 5-fold CV (based on ECG+Actigraphy with FS(\#Feat.15)).  
We can see features 1) very low frequency (75th percentile), 2) range NNI (maximum), 3) Mod. Cardiac Symp.IdNx (50th percentile), and 4) high frequency (std) have contributed the most to the general performance, which might give some insights to clinicians and practitioners.  
More details of these features can be found at Table \ref{tab:featureHRV}.

\subsubsection{Deep Learning Models}
\begin{figure}[]
    \begin{subfigure}[b]{0.25\textwidth}
        \centering
        \includegraphics[width=0.95\textwidth]{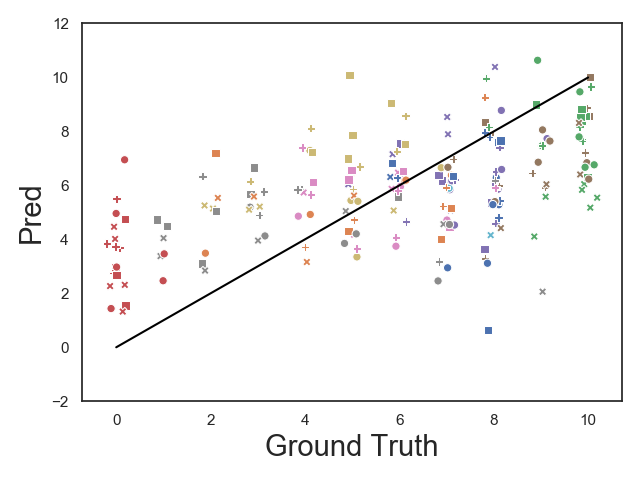}
        \caption{Linear Regression (corr: 0.53)}
    \end{subfigure}%
    \begin{subfigure}[b]{0.25\textwidth}
        \centering
        \includegraphics[width=0.95\textwidth]{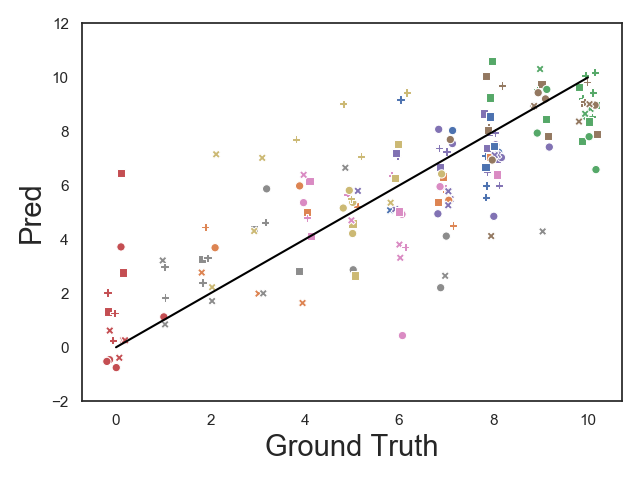}
        \caption{LSTM-CSA (corr: 0.81)}
    \end{subfigure}%
    \caption{Scatter Graphs (ECG+Actigraphy; 5-fold CV)}
    \label{fig:pred_gt_scatter}
\end{figure}

\hfill\\
For all LSTM models, we used one hidden layer with $D_l = 128$ and set batch size to 1; for LSTM-SA/LSTM-CSA, we set $D_a = 128$.
Main results of the LSTM models were reported in Table \ref{tab:results_DL}, from which we can see that LSTM-CSA achieved the better results than other models irrespective of modalities. 
In contrast to linear models, LSTM-CSA can exploit useful information from Actigraphy, making the combined modality with the best results (than ECG only).
Based on combined modality, we also presented the scatter graphs of linear model and LSTM-CSA in Fig. \ref{fig:pred_gt_scatter}, and we can see LSTM-CSA has a larger correlation coefficient (between ground truth and prediction).
We also visualised the attentions of LSTM-SA, and LSTM-CSA in Fig. \ref{fig:att_weights} in Appendix for a sequence with some missing data, and we can see the consistency nature (i.e.,smoothness) of LSTM-CSA's attention, in contrast to the non-smooth attention from LSTM-SA.

\begin{table}[]
\caption{Results of deep learning models}
\begin{tabular}{|l|l|l|l|}
\hline
Modalities                                                                    & Models & MAE             & RMSE           \\ \hline
\multirow{3}{*}{Actigraphy}                                                & LSTM                         & 2.50$\pm$0.33       & 2.93$\pm$0.28      \\ \cline{2-4} 
                                                                           & LSTM-SA               & 2.32$\pm$0.36       & 2.78$\pm$0.29      \\ \cline{2-4} 
                                                                           & LSTM-CSA                         & 2.23$\pm$0.27       & 2.78$\pm$0.25      \\ \hline
\multirow{3}{*}{ECG}                                                       & LSTM                         & 2.03$\pm$0.31       & 2.58$\pm$0.28      \\ \cline{2-4} 
                                                                           & LSTM-SA               & 1.62$\pm$0.37       & 2.09$\pm$0.42      \\ \cline{2-4} 
                                                                           & LSTM-CSA                         & \textbf{1.43$\pm$0.30}       & \textbf{1.80$\pm$0.40}      \\ \hline
\multirow{3}{*}{\begin{tabular}[c]{@{}l@{}}ECG+\\ Actigraphy\end{tabular}} & LSTM                         & 1.96$\pm$0.17       & 2.53$\pm$0.26      \\ \cline{2-4} 
                                                                           & LSTM-SA               & 1.75$\pm$0.28       & 2.15$\pm$0.37      \\ \cline{2-4} 
                                                                           & LSTM-CSA                         & \textbf{1.39$\pm$0.25}       & \textbf{1.78$\pm$0.37}      \\ \hline
\end{tabular}
\label{tab:results_DL}
\end{table}

\subsection{Limitations and Discussion}
\begin{figure}[]
 \centering
 \includegraphics[width=0.45\columnwidth]{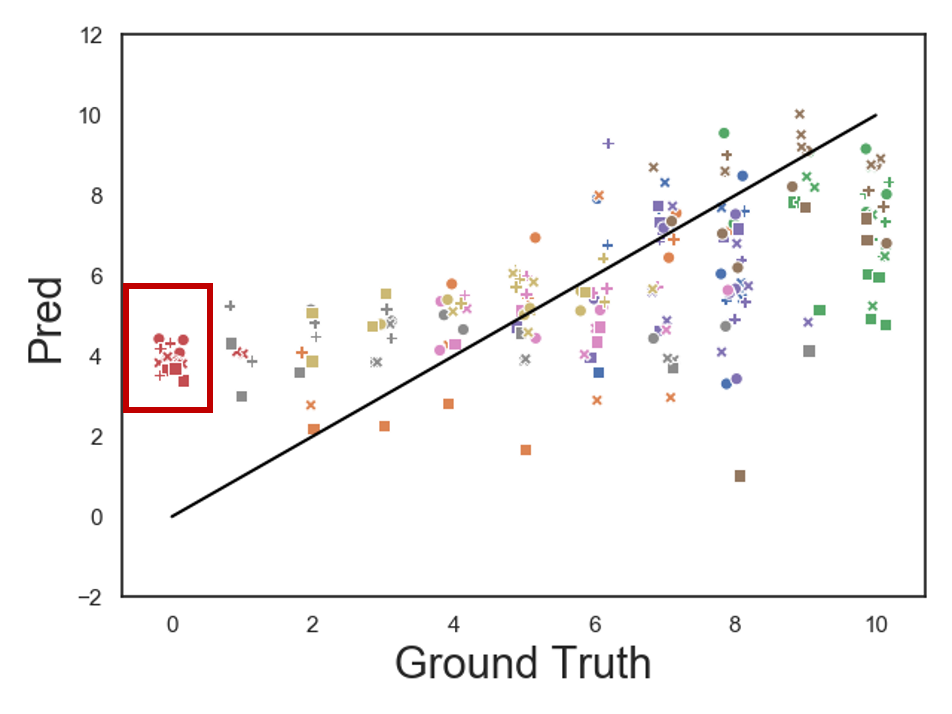}  
 
 \caption{LSTM-CSA with ECG+Actigraphy using LOSO-CV (corr:0.63); different colours denoting different subjects}
 \label{fig:LOSO-CV}
\end{figure}

Although we saw promising results of LSTM-CSA on the combined modality, it was based on 5-fold CV. 
For practical applications, we also performed leave-one-subject-out-cross-validation(LOSO-CV), and the results were shown in Fig. \ref{fig:LOSO-CV}, from which we can see there is a significant performance drop (e.g., correlation coefficient from 0.81 to 0.63).   
One of the major reason is the small subject number in this study. 
For example, \textbf{only one} subject reported 'No fatigue' (with score 0 in questionnaire, see the red box in Fig. \ref{fig:LOSO-CV}), making the trained model (on other subjects) hardly generalise to that subject. Such overfitting problem can be reduced by using larger dataset.

In this work, although deep learning solution can provide better results, they are less transparent than the interpretable solution, from which we can list the key human-understandable features.
However, interpretable solution relies heavily on feature engineering. 
For example, we used the simplest features from Actigraphy, and these features played a negative role, making the combining effect of ECG+Actigraphy worse than ECG only.
On the other hand, deep learning is a data-driven approach and it can learn the useful information from Actigraphy, making combined modality with the best results.
Nevertheless, both solutions have their own advantages and they can be used together.  
For example, deep learning can be used as a prototype tool to guide feature engineering for interpretable solution, which will be explored in the future.


\section{Conclusion and Future Work}
To develop an automated fatigue assessment system, in this work we introduced a pipeline from data collection, data preprocessing, feature engineering to interpretable solution and deep learning solution. 
Both solutions were evaluated on the collected dataset, and some promising results were achieved.

This work is a pilot study of a larger project, where 120 subjects will be recruited, based on which the proposed solutions' generalisation capability will be further evaluated. 
Next, we will also explore 1) how to use deep learning as a prototype tool to guide feature engineering for interpretable solution; 2) how to further improve the generalisation capability of the deep learning solution (e.g., LSTM-CSA), e.g., by using LSTM ensemble \cite{LSTM_ensemble}.

\begin{acks}
We'd like to thank Dr. Valentin Hamy and Dr. Luis Garcia-Gancedo from GSK for the valuable discussion.
This work is supported by the Newcastle NIHR Biomedical Research Centre; and also funded by the Innovative Medicines Initiative 2 Joint Undertaking (JU) (IDEA-FAST grant agreement No. 853981) receiving support from the EU's H2020 research and innovation program and EFPIA.

\end{acks}

\bibliographystyle{ACM-Reference-Format}
\bibliography{sample-base}

\appendix
\section{Appendix}
\begin{table}[H]
\scriptsize
\caption{Demographic information for Participants.}
\begin{tabular}{|l|l|l|l|l|l|l|l|l|l|}
\hline
Participant & 1 & 2 & 3 & 4 & 5 & 6 & 7 & 8 & 9 \\ \hline
Gender(M/F)      & F  & F  & F  & M    & M    & F  & M    & F  & F  \\ \hline
Age(y)         & 63      & 81      & 36      & 70      & 62      & 54      & 74      & 64      & 44      \\ \hline
Height(m)   & 1.7     & 1.55    & 1.7     & 1.8     & 1.75    & 1.73    & 1.73    & 1.6     & 1.6     \\ \hline
Weight(kg) & 76.6    & 70.3    & 70.3    & 71.2    & 96.2    & 70.3    & 74.8    & 59      & 56.7    \\ \hline
\end{tabular}
\label{tab:pinfo}
\end{table}

\begin{figure}[H]
\centering
\includegraphics[width=0.9\columnwidth]{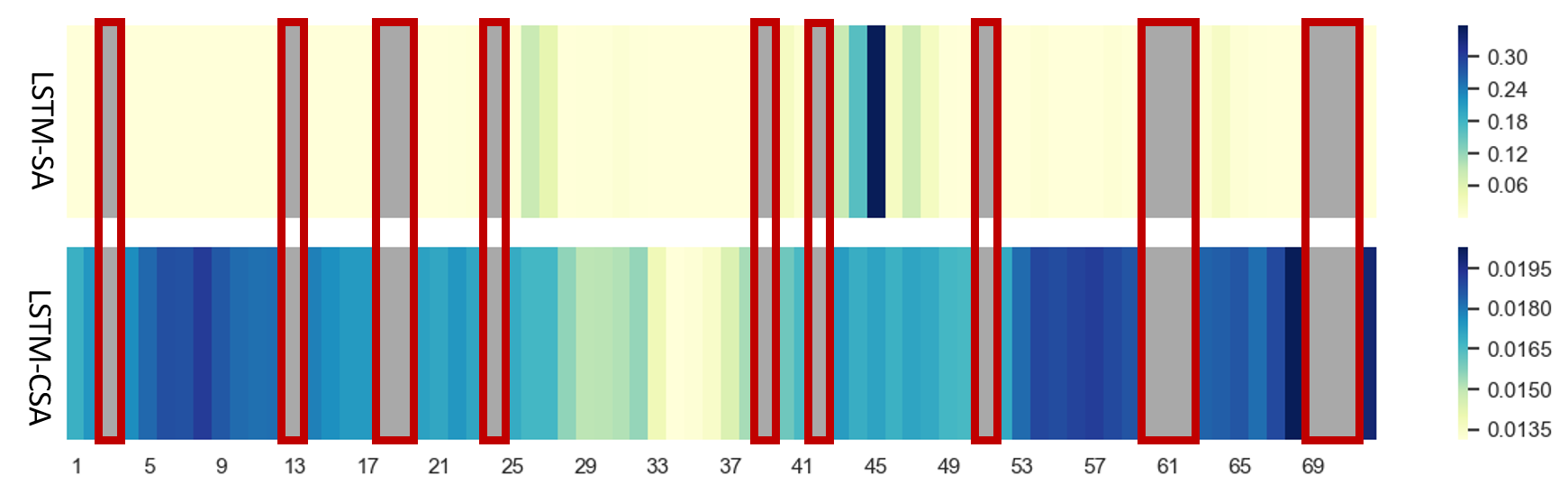}  
\caption{Attention visualization for LSTM-SA (top) and LSTM-CSA (bottom); Red boxes denoting the missing data.}
\label{fig:att_weights}
\end{figure}


\end{document}